\documentclass{article}

\usepackage{arxiv}

\usepackage[utf8]{inputenc} 
\usepackage[T1]{fontenc}    
\usepackage{hyperref}       
\usepackage{url}            
\usepackage{booktabs}       
\usepackage{amsfonts}       
\usepackage{nicefrac}       
\usepackage{microtype}      
\usepackage{lipsum}		
\usepackage{graphicx}
\usepackage{natbib}
\usepackage{doi}
\usepackage{eurosym}
\usepackage{rotating} 
\usepackage{multirow} 
\usepackage{rotating}
\usepackage{array}

\title{Explainable artificial intelligence model for identifying Market Value in Professional Soccer Players}

\author{ \href{https://orcid.org/0009-0003-3150-1254}{\includegraphics[scale=0.06]{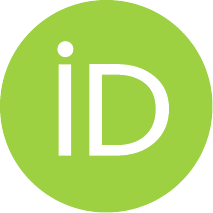}\hspace{1mm}Chunyang Huang}\thanks{Use footnote for providing further
		information about author (webpage, alternative
		address)---\emph{not} for acknowledging funding agencies.} \\
	Zhili College\\
	Tsinghua University\\
    Beijing 100084, China\\
	\texttt{cy-huang20@mails.tsinghua.edu.cn} \\
	\And
	\href{https://orcid.org/0000-0002-0543-7135}
        {\includegraphics[scale=0.06]{orcid.pdf}\hspace{1mm}Shaoliang Zhang} \\
	Research Centre for Athletic Performance and Data Science \& Division of Sports Science and Physical Education\\
	Tsinghua University\\
    Beijing 100084, China\\
	\texttt{zslinef@mail.tinghua.edu.cn} \\
}

\renewcommand{\shorttitle}{\textit{arXiv} Template}

\hypersetup{
pdftitle={A template for the arxiv style},
pdfsubject={q-bio.NC, q-bio.QM},
pdfauthor={Chunyang Huang, Shaoliang Zhang},
pdfkeywords={Player Market Value Prediction, Shapley Additive Explanations (SHAP), Ensemble Machine Learning Models, Soccer or football}, 
}

\begin{document}
\maketitle
\begin{abstract}
This study introduces an advanced machine learning method for predicting soccer players' market values, combining ensemble models and the Shapley Additive Explanations (SHAP) for interpretability. Utilizing data from about 12,000 players from Sofifa, the Boruta algorithm streamlined feature selection. The Gradient Boosting Decision Tree (GBDT) model excelled in predictive accuracy, with an R-squared of 0.901 and a Root Mean Squared Error (RMSE) of 3,221,632.175. Player attributes in skills, fitness, and cognitive areas significantly influenced market value. These insights aid sports industry stakeholders in player valuation. However, the study has limitations, like underestimating superstar players' values and needing larger datasets. Future research directions include enhancing the model's applicability and exploring value prediction in various contexts.
\end{abstract}


\section{Introduction}
Soccer, often referred to as 'the beautiful game', holds preeminence as a global sport, captivating a diverse audience across various cultural and geographical landscapes. Its allure transcends the physical boundaries of the playing field, significantly bolstering a multi-billion-dollar economic industry. This industry is characterized by varied revenue streams, including broadcasting rights, ticket sales, merchandising, sponsorships, and notably, the transfer market. The latter is subject to extensive financial scrutiny, underscoring its economic impact  \citep{dobson_economics_2001}.

The process of estimating a player's market value is a critical aspect of the economic operations within football, shaping the sport's financial landscape. These valuations are crucial during transfer negotiations, reflecting a player’s potential impact on a team's performance and economic success. High-profile transfers, involving substantial financial investments, not only affect a club's prestige but also its appeal, making accurate player valuation essential for maximizing financial returns  \citep{wolfers_prediction_2004}. Furthermore, a player's market value influences wage policies within clubs, with top-tier players commanding higher salaries. This, in turn, affects the financial stability and budgetary planning of the clubs. Additionally, the collective market value of players plays a pivotal role in a club’s overall valuation, particularly during ownership transitions or financial negotiations, highlighting the significance of precise player valuation in the strategic fiscal management of football clubs.

In the contemporary era, dominated by 'big data', analytical methodologies have become indispensable in sports, particularly in the context of player market valuation. These methodologies inform decision-making processes related to player recruitment and selection. The emergence of sophisticated online platforms such as Sofifa, WhoScored, and Transfermarkt, offering comprehensive player data, has transformed this area. The integration of machine learning algorithms with these data sources aims to enhance the precision of player market valuations \citep{baboota_predictive_2019}. For instance, \cite{mustafa_a_al-asadi_predict_2022}analyzed FIFA 20 video game data from Sofifa.com, employing four regression models—linear regression, multiple linear regression, decision trees, and random forests—to evaluate players' market values. This study demonstrated that the random forest model outperformed traditional statistical models in predicting players’ market values, with players’ potential, international reputation, age, and height being key determinants of individual market value. Similarly, \cite{mchale_estimating_2023} analyzed transfer details and crowd-sourced player ratings from transfermarkt.com and sofifa.com over nine seasons. The addition of advanced player rating systems from sofifa.com, beyond player performance profiles, significantly improved the predictive accuracy of the models, with XGBoost showing a substantial increase in predictive accuracy compared to linear regression models. Moreover, \cite{yang_predicting_2022}compiled a longitudinal transfer dataset from the top five European leagues over 14 seasons from transfermarkt.de. The study employed generalized, quantile additive models, and random forests to predict players' market values, concluding that machine learning models, particularly random forests, are superior in evaluating market values. The study also identified non-linear predictors of transfer fees, such as buying-club expenditure and selling-club income, as having significant impacts on transfer fees, overshadowing players’ anthropometric characteristics and technical performance. Despite the advancements in machine learning techniques that improve predictive accuracy, the challenge of attaining clear interpretability among various models at both the global and local scales persists.

In light of these developments, this study adopts ensemble machine learning models, concentrating on the most important factors influencing athletic performance. Utilizing the SHapley Additive exPlanations (SHAP) method, it achieves transparent and comprehensive interpretive visualizations from both local and global perspectives. These insights identify the primary features affecting athlete performance, offering a detailed quantitative analysis of individual player metrics that contribute to their market value.

\section{Methods}
\subsection{Sample}
\label{subsection: Sample}
Building upon the findings of previous research, this study conducts an in-depth analysis of the dataset available on Sofifa.com, a widely recognized resource among FIFA football manager enthusiasts. The website offers extensive data, including but not limited to, nuanced player ratings, team compositions, and a variety of other statistics pertinent to the game. Further, it provides detailed information on aspects such as player positions and their preferred playing foot.

In the course of our research, data was systematically extracted relating to approximately 12,000 players from Sofifa.com, as recorded on January 5, 2023. This dataset is comprehensive, encompassing a multitude of attributes including players' names, market values, wages, overall ratings, potential, and an additional 34 features. Among these, 29 features are applicable to outfield players, while 5 are uniquely relevant to goalkeepers. The table \ref{table: feature description} in Appendix A details these features, offering a thorough overview essential for our analysis

\begin{figure}
	\centering
        \includegraphics[width=0.8\textwidth]{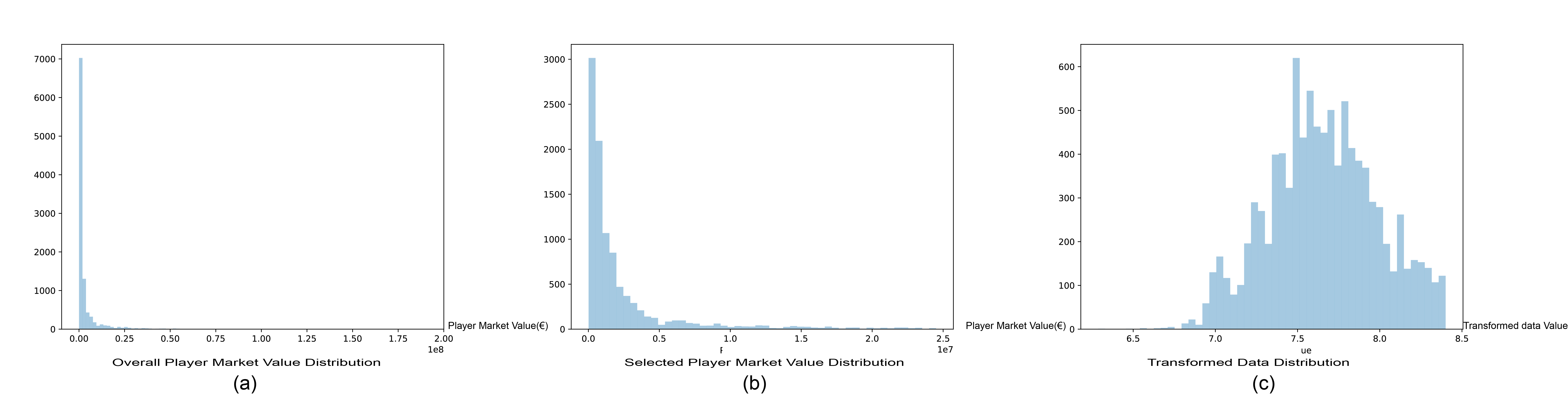}
	\caption{The distribution of origin value(1a), deducted value(1b) and transformed data(1c)}
	\label{fig:data description}
\end{figure}

In the initial phase of our methodology, the dataset underwent a rigorous cleansing process to rectify any instances of missing values and to categorize the data into two primary classifications: outfield players and goalkeepers. The dataset exhibited a wide range of player values, stretching from \euro 15,000 to \euro 190 million. Figure \ref{fig:data description} a graphically represents the distribution of these player values.

Upon examining the distribution in Figure \ref{fig:data description} a, a notable aggregation near the lower end of the value spectrum was observed. This pattern suggests that a limited number of high-value players disproportionately influence the extension of the horizontal axis in the graph. This distributional characteristic aligns with the concept of player popularity and the 'superstar' effect, as discussed in seminal works by \cite{adler_stardom_1985},  \cite{franck_talent_2012} and \cite{rosen_economics_1981}. However, these elements are extraneous to the performance-centric focus of our analysis, as delineated by \cite{muller_beyond_2017}. Consequently, data points representing player values exceeding \euro 25 million, constituting approximately 3\% of the dataset, were excluded, as illustrated in Figure \ref{fig:data description}b.

The revised distribution, as depicted in Figure \ref{fig:data description}b, revealed a significant skewness towards the lower value range. To rectify this skewness and foster a more symmetrical distribution conducive to effective statistical modeling, we employed the Box-Cox transformation, a technique well-established in statistical literature \citep{box_analysis_1964,osborne_improving_2019}. This transformation was implemented utilizing the Scipy library within the Python programming environment \citep{virtanen_scipy_2020}. The outcome of this transformation, which significantly improved the symmetry of the data distribution, is displayed in Figure \ref{fig:data description}c.

\subsection{Feature selection}
The initial step in our analytical process involves a critical task: feature selection. Our dataset includes up to 29 attributes related to football players' performance. Within this set, certain features may contribute minimally to the predictive accuracy of our model, presenting several challenges. Firstly, an excess of features can lead to increased computational time and escalate the computational power requirements. Secondly, the elimination of non-essential features, as suggested by \cite{kohavi_wrappers_1997}, is likely to enhance the precision of our machine learning model.

To address these concerns, this study employs the Boruta algorithm for feature selection. This algorithm offers a comprehensive solution to issues extending beyond minimal-optimal feature sets \citep{daoud_multicollinearity_2017, kursa_feature_2010, nilsson_consistent_2007}. The Boruta algorithm, a wrapper built around the random forest classification algorithm, aims to identify all potentially significant features in the dataset, including those that might otherwise be overlooked. It operates through an iterative process, comparing the importance of original features with shadow features - which are randomized permutations of the original set - to ascertain their relative significance.

\subsection{Model selection}
In the pursuit of developing an optimal predictive model for assessing the market value of football players, this study rigorously evaluates a variety of learning algorithms. The algorithms chosen for this analysis include Adaboost \citep{freund_decision-theoretic_1997}, LightGBM \citep{ke_lightgbm_2017}, \citep{ho_random_1995}, Gradient Boosting Decision Tree (GBDT)  \citep{friedman_greedy_2001}, CatBoost (Dorogush et al., 2018) \citep{dorogush_catboost_2018}, and XGBoost \citep{chen_xgboost_2016}. These were selected for their relevance and potential efficacy in forming the foundational structure of the models.

The methodological approach centers on ensemble learning algorithms, a class of meta-algorithms that synthesize the methodologies of individual models to construct a comprehensive predictive framework \citep{webb_multistrategy_2004}. The rationale behind selecting ensemble modeling lies in its multiple benefits, which include diminishing variance and bias while concurrently enhancing the overall predictive accuracy of the model \citep{sagi_ensemble_2018}. Ensemble learning, by amalgamating predictions from multiple models, inherently demonstrates a capacity for superior performance in comparison to singular model approaches.

\subsection{Model development and tuning}
In the construction of the predictive model, the dataset underwent a randomized spliting process: 80\% was allocated for training and model validation, while the remaining 20\% was designated for testing purposes. To maintain the analytical integrity and avert any potential data leakage, a rigorous methodology was employed. This method entailed conducting both imputation and feature selection exclusively on the training set prior to undertaking any evaluative measures. Such a procedural approach was integral in ensuring that the test dataset, reserved solely for the ultimate evaluation of classifier efficacy, remained devoid of any influences that could introduce bias into the model's performance assessment.

To optimize the hyperparameters of each ensemble learning model, we employed a 5-fold cross-validation technique coupled with Grid Search. This rigorous process allowed us to effectively tune the models and obtain reliable and unbiased performance estimates \citep{yarkoni_choosing_2017}. The training dataset was initially partitioned into ten distinct subsets of equal size through random assignment. Out of these subsets, nine were used for training the model, while the remaining subset served the purpose of validation. This iterative procedure was repeated for all ten feasible combinations. Ultimately, a prognostic model was constructed by leveraging discerning features and refining hyperparameters for optimal performance \citep{raschka_model_2018}.
  
\subsection{Model evaluation}
The evaluation of various machine learning algorithms constitutes a critical component of this study. In this context, a comprehensive assessment was conducted utilizing multiple metrics to gauge the accuracy of predictions pertaining to player market value. Key among these metrics are the R-squared value, which quantifies the proportion of variance in the dependent variable that is predictable from the independent variables, and the Root Mean Squared Error (RMSE), which provides a measure of the average magnitude of the prediction errors. By employing both R-squared and RMSE, we aim to offer a comprehensive and multifaceted evaluation of our model's performance.
\subsection{Model interpretation}
In the realm of machine learning, models are often characterized as 'black boxes,' presenting challenges in interpreting the underlying mechanisms driving their predictive accuracy, particularly in the context of evaluating player market values \citep{linardatos_explainable_2021}. To surmount these interpretability hurdles, Lundberg and Lee introduced the Shapley Additive exPlanations (SHAP) approach. This method employs Shapley values, a well-established metric for assessing feature importance, to elucidate the outputs of any machine learning model. SHAP facilitates both global and local interpretability, enabling an analysis of how individual input characteristics contribute positively or negatively to the predictive outcomes.

For global interpretability, this study utilizes the SHAP beeswarm plot and feature importance measures. The beeswarm plot ranks features based on their influence within the predictive model, with the y-axis representing the features and the x-axis indicating the respective SHAP values. Each feature is depicted by rows of colored dots, where red indicates high values and blue denotes low values, thus providing a clear visual representation of the impact each feature has on the model’s output \citep{rommers_machine_2020}.

In the pursuit of local interpretability, the SHAP force plot was implemented to specifically elucidate the predicted market value of individual players. This analytical tool clarifies the contribution of each feature to the prediction for a given sample, graphically illustrating the trajectory from the base value (average predicted value) to the final predicted outcome \citep{hiabu_unifying_2023}. Features that positively influence the prediction, leading to higher values, are represented in red, while those that negatively impact the prediction, resulting in lower values, are shown in blue. This visualization offers a granular perspective of the model's predictions, underscoring the significance of each feature in ascertaining the market value of the players under consideration.

Additionally, to assess the SHAP value matrix for each feature, the Partial Dependence Plot (PDP) was generated. The PDP elucidates the marginal effect of a specific feature on the predicted outcome by averaging out the effects of all other features. It essentially portrays the average prediction for varying values of the targeted feature, thereby isolating its influence from other contributing factors.

\section{Result}
\begin{figure}
	\centering
        \includegraphics[width=0.8\textwidth]{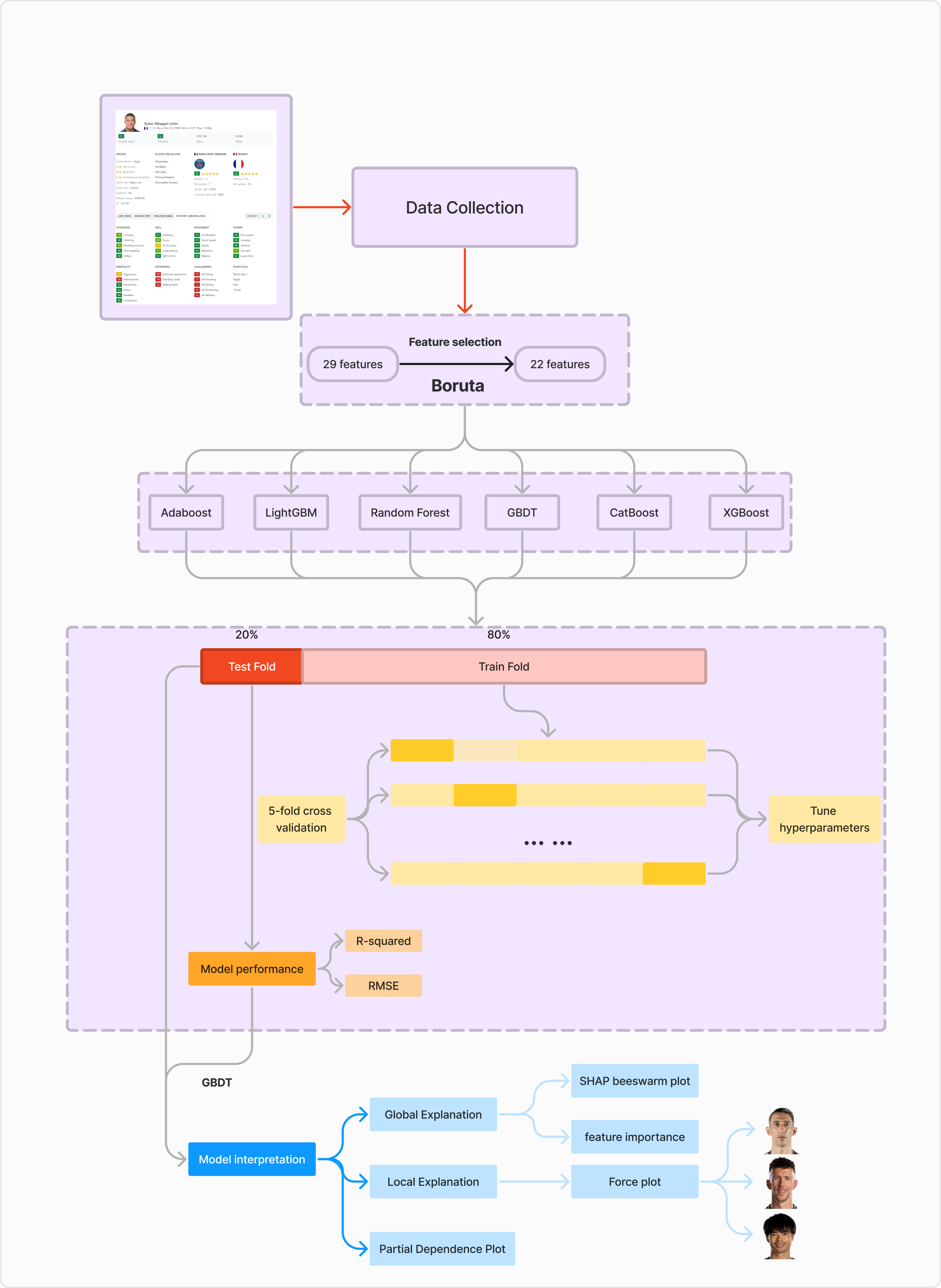}
	\caption{Flow chart of the overall study design with data collection, feature screening, model development and validation, and model evaluation and interpretation}
	\label{fig:workflow}
\end{figure}

The methodological framework of this study is depicted in Figure \ref{fig:workflow}, illustrating the comprehensive study design inclusive of data collection, feature screening, model development and validation, and model evaluation and interpretation.
\subsection{Feature selection}
\begin{figure}
	\centering
	\includegraphics[width=0.8\textwidth]{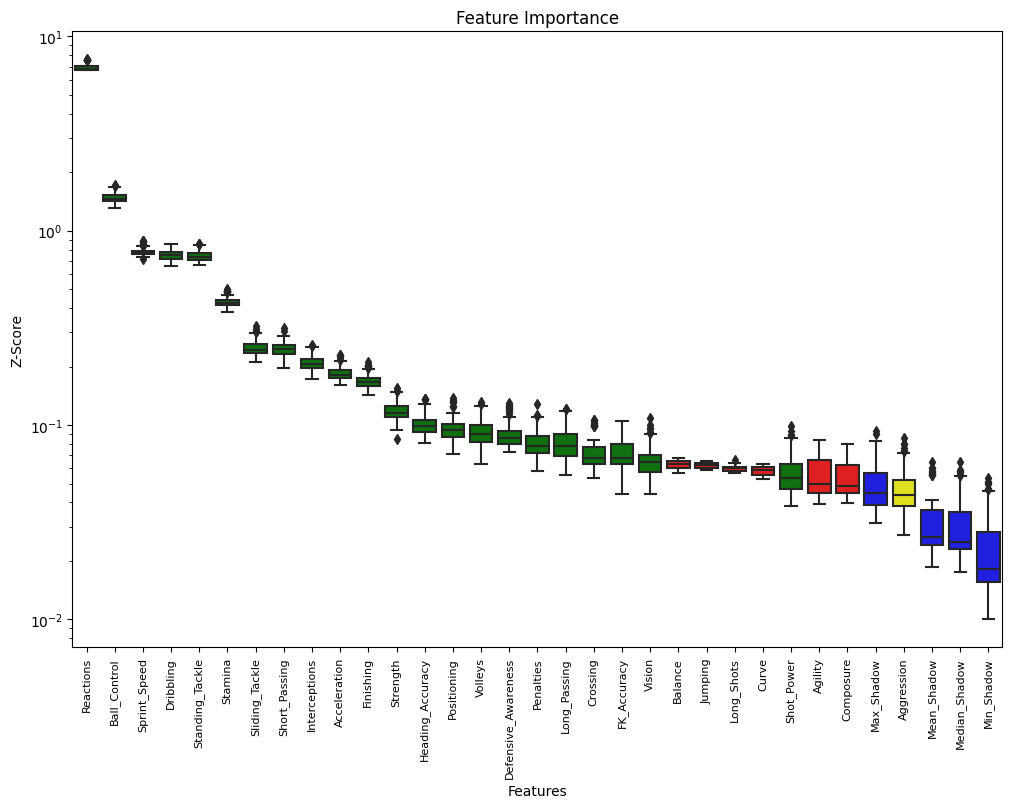}
	\caption{Feature selection using the Boruta algorithm for key features}
	\label{fig:boruta}
\end{figure}

In the initial phase of feature selection, we commenced with a set of 29 features. Utilization of the Boruta algorithm, implemented through the BorutaShap package in Python, allowed for a reduction in the feature set to 22, as indicated by the green bars in Figure \ref{fig:boruta}. Acceleration, Heading accuracy, Defensive awareness, Vision, Volleys, Sprint speed, Long passing, Positioning, Standing tackle, Dribbling, FK accuracy, Short passing, Interceptions, Penalties, Finishing, Reactions, Ball control, Stamina, Crossing, Strength, Shot power, Sliding tackle.

\subsection{Model development and evaluation}
The outcomes of the cross-validation analysis and evaluation on test set are detailed in Table \ref{table: cross-validation}. Within the ensemble of six learning algorithms evaluated, the Gradient Boosting Decision Tree (GBDT) model demonstrated superior performance, achieving the highest R-Squared value of 0.889. The CatBoost model was a close second with an R-Squared of 0.887, followed by LightGBM with 0.885. The Random Forest and XGBoost models obtained R-Squared values of 0.877 and 0.861, respectively, while the AdaBoost model had the lowest R-Squared at 0.773. Regarding the Root Mean Squared Error (RMSE), the GBDT model surpassed all counterparts, recording an RMSE of 3,060,228.569. The subsequent models, CatBoost (4,715,039.662), LightGBM (3,249,280.179), Random Forest (3505068.8371), XGBoost (3,320,149.832), and AdaBoost (4,442,839.041), followed in ascending order of RMSE values. Notably, within the test set, the GBDT model sustained its advantage, achieving the highest R-squared value of 0.901 and the lowest RMSE of 3,221,632.175, indicative of its robustness in predicting player market values.

\begin{table}[!ht]
    \centering
    \caption{The cross-validation analysis and evaluation on test set}
    \begin{tabular}{lllll}
    \hline
    \multirow{2}{*}{Model Name} & \multicolumn{2}{c}{Cross Validation} & \multicolumn{2}{c}{Test Set} \\ \cline{2-5} 
    & Mean R\textsuperscript{2} & Mean RMSE & Mean R\textsuperscript{2} & Mean RMSE \\ \hline
    AdaBoost & 0.764 & 4442839.041 & 0.752 & 4657341.125 \\ 
    GBDT & 0.878 & 3221632.175 & 0.901 & 3221632.175 \\ 
    LightGBM & 0.877 & 3249280.179 & 0.886 & 3157342.577 \\ 
    Random Forest & 0.856 & 3505068.837 & 0.856 & 3547133.366 \\ 
    XGBoost & 0.871 & 3320149.832 & 0.888 & 3127608.099 \\ 
    CatBoost & 0.742 & 4715039.662 & 0.742 & 4715039.662 \\ \hline
    \end{tabular}
    \label{table: cross-validation}
\end{table}

\subsection{Global and local interpretation of the ML Model}
\begin{figure}
	\centering
	\includegraphics[width=0.8\textwidth]{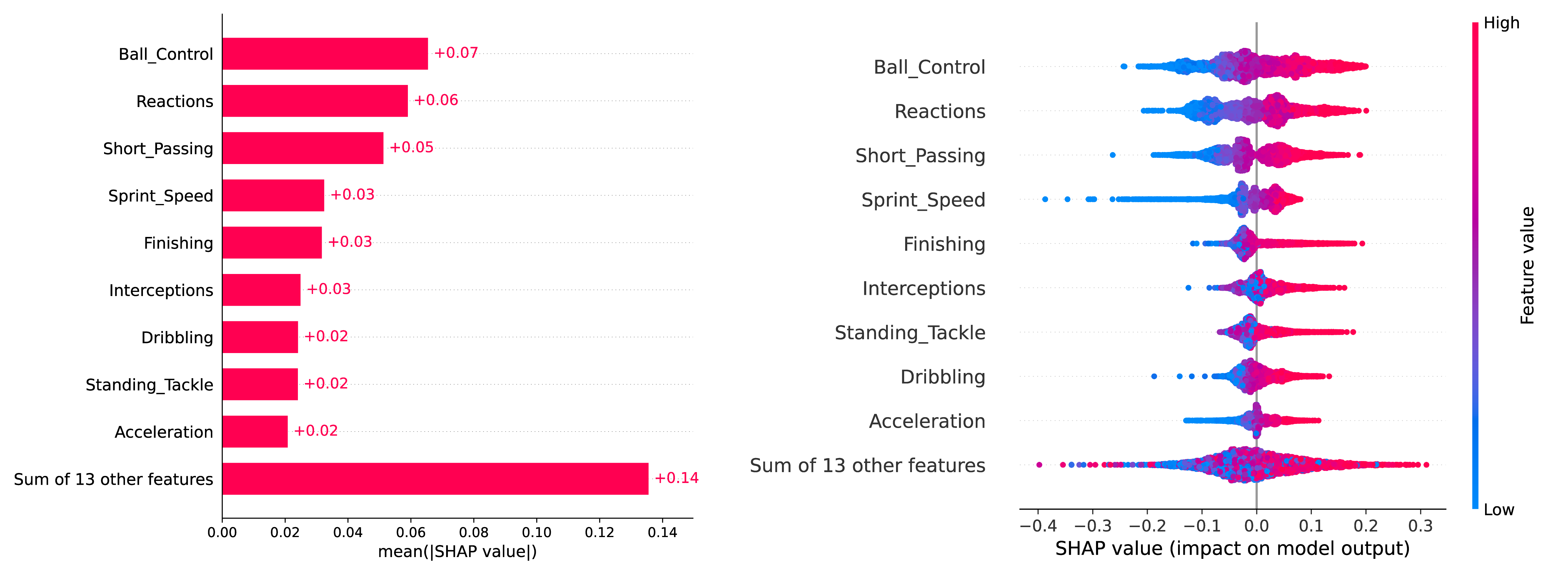}
	\caption{The GBDT model’s interpretation. The importance ranking of the different variables according to the mean (|SHAP value|) (a); The importance ranking of different risk factors with stability and interpretation using the optimal model (b). The higher SHAP value of a feature is given, the higher market value of players would have. The red part in feature value represents higher value.}
	\label{fig:global explanation}
\end{figure}

In the study, the SHAP (Shapley Additive Explanations) beeswarm plot and feature importance for the Gradient Boosted Decision Trees (GBDT) model, as illustrated in Figure \ref{fig:global explanation}, were employed to identify the features with the most significant influence on the prediction model. The analysis revealed that nine variables—Ball control, Reactions, Short passing, Sprint speed, Finishing, Interceptions, Dribbling, Sliding Tackle, and Acceleration—held the most substantial predictive power, significantly impacting a player’s market value.

\begin{figure}
	\centering
        \includegraphics[width=0.8\textwidth]{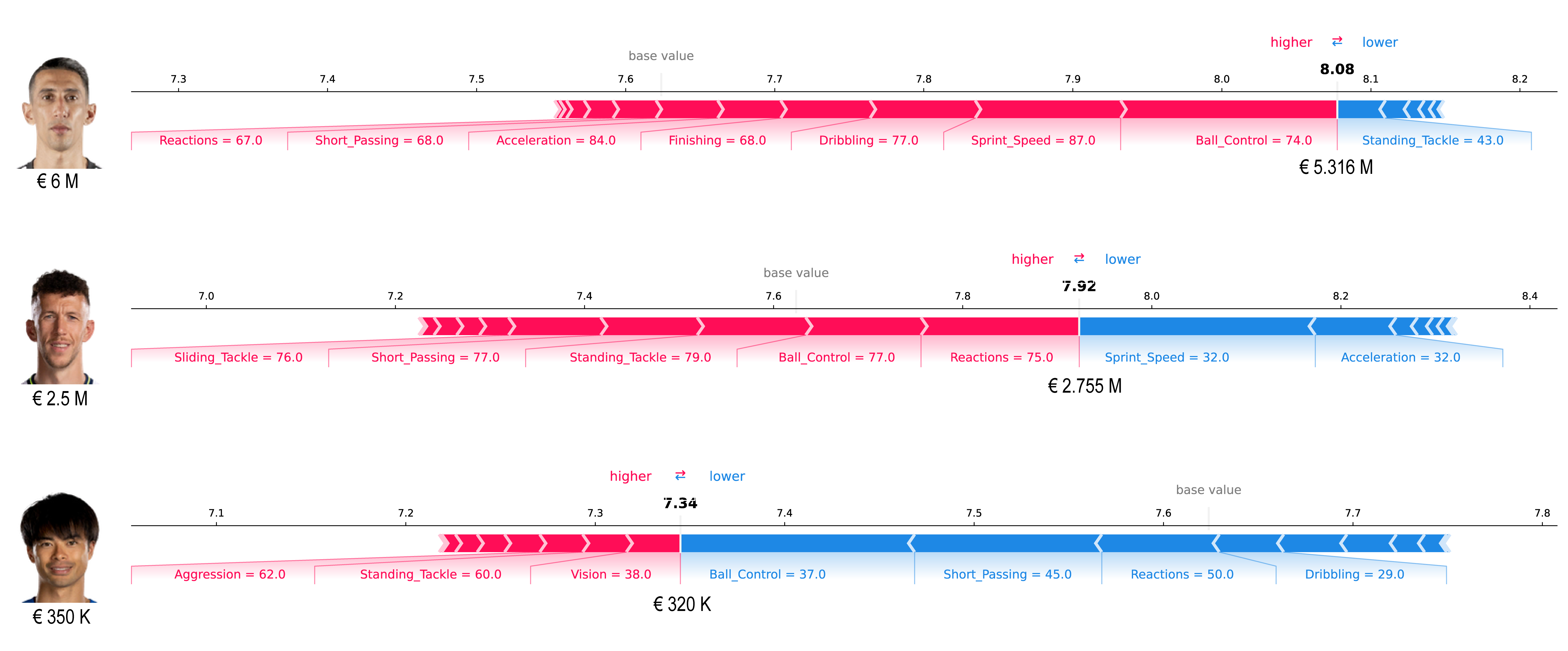}
	\caption{SHAP force plot for predicting players’ market value. (A) SHAP forces plot to correctly predict market value of Ángel Fabián. (B) SHAP forces plot to correctly predict market value of Ivan Perišić (C) SHAP force plot correctly predict market value of Teruki Miyamoto. Red features mean pushing the prediction higher market value and blue features pushing the prediction lower market value.}
	\label{fig:local explanation}
\end{figure}

This detailed investigation into the predicted market values for players such as Ángel Fabián, Ivan Perišić, and Teruki Miyamoto, as illustrated in Figure \ref{fig:local explanation}, highlights the practical applicability and accuracy of the Gradient Boosted Decision Trees model in real-world scenarios. For the first player, Ángel Fabián, the predicted market value, post Box-Cox transformation, was approximately \euro 6 Million. This prediction, derived from a transformed value of 8.08, closely aligns with his actual market value of \euro 5.31 million. In the case of Ivan Perišić, the predicted market value following the Box-Cox transformation was approximately \euro 2.5 Million, based on a transformed prediction of 7.92. This estimate is near the real market value of \euro 2.75 million. Finally, for Teruki Miyamoto, the model estimated a market value of approximately \euro 3.20 Million after the transformation, calculated from a transformed prediction of 7.34, which is in close proximity to the actual market value of \euro 3.50 million.

\subsection{Partial Dependence Plot}
\begin{figure}
	\centering
	\includegraphics[width=0.8\textwidth]{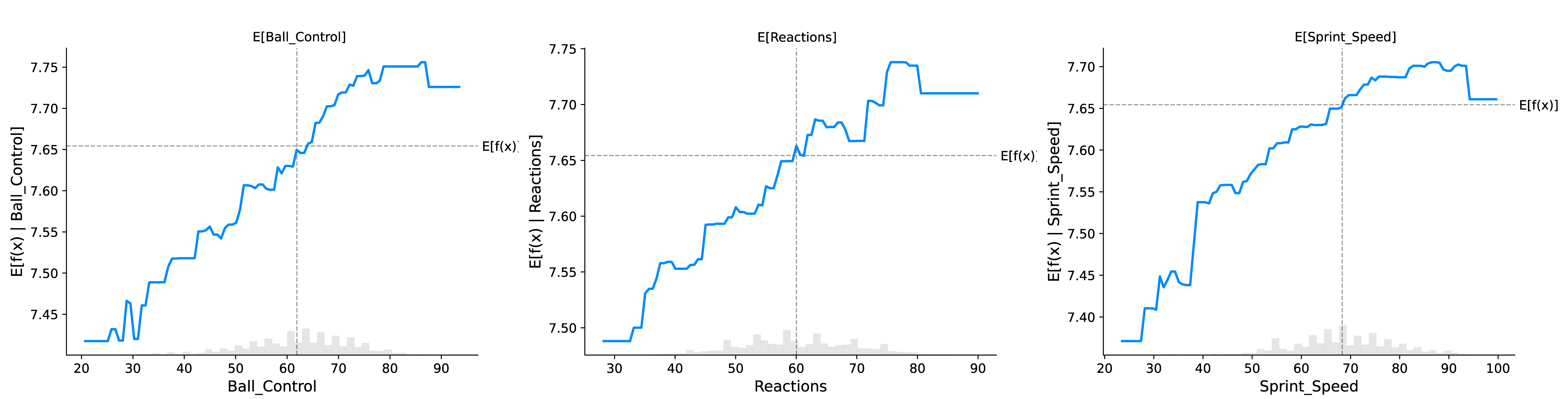}
	\caption{partial dependence plots of top features in skill, cognitive, and fitness dimensions}
	\label{fig:pdp}
\end{figure}

The Partial Dependence Plots (PDP) for these features are presented in Figure \ref{fig:pdp}, offering a detailed examination of their marginal effects on the predicted outcomes. It was observed that the SHAP values of Ball Control, Reactions, and Sprint Speed generally exhibited an increasing trend in correspondence with the escalation of the respective market values. This trend underscores a direct relationship between these specific player attributes and their market valuation, thereby highlighting their critical importance in the predictive model.

\section{Discussion}
The aim of this study is to ensemble machine learning models, concentrating on the most important factors influencing athletic performance. Utilizing the SHapley Additive exPlanations (SHAP) method, it achieves transparent and comprehensive interpretive visualizations from both local and global perspectives. In previous research, methodologies such as static statistical methods, linear regression, and conventional machine learning approaches have demonstrated limited effectiveness in achieving outstanding predictive performance. Moreover, these methods often fall short in providing in-depth insights into each feature's contribution, thereby limiting the explainability of the prediction model. In contrast, the present study, utilizing data scrupulously gathered from the Sofifa website, employs an ensemble machine learning model. This model not only achieves exceptional accuracy in predicting market values but also leverages Shapley Additive Explanations (SHAP) to gain both global and local insights into all features initially selected by the Boruta method. The study successfully identifies key features that are crucial based on their global explanation and their response tendencies to varying feature values.

The evaluation of feature influences is of paramount importance in the estimation of players' market and transfer values. Traditionally, club managers have considered three key dimensions in assessing players and their market worth: skills, fitness, and cognition. Our study reveals distinct insights within these dimensions. Specifically, in the skills dimension, Ball Control, Short Passing, Finishing, Interceptions, Dribbling, and Tackling emerge as the most influential factors. In the fitness dimension, Sprint Speed and Acceleration are identified as having the greatest impact. Notably, within the cognition dimension, Reactions are found to hold the most significant position. These findings regarding the impact of various features can guide managers to focus on the most influential aspects during the player valuation process. By understanding the relative importance of these attributes, club managers can make more informed decisions, leading to more accurate and effective player assessments and strategic planning in the realm of player acquisitions and transfers. 

The exemplary performance of our model in evaluation phases suggests that the predicted market values closely align with actual figures, thereby validating the effectiveness of our prediction approach. However, it is crucial to highlight that this enhanced prediction performance is achieved at the cost of some interpretability of the model output. In our methodology, we employed the Box-Cox transformation to convert the original exponential distribution into a single-peak distribution. This transformation involved converting the actual market values into transformed data, which facilitated an increase in prediction accuracy. Consequently, the direct output of our model is in the form of predicted transformed data. This necessitates an inverse Box-Cox transformation to revert these predictions back to the estimated market values. Furthermore, the SHAP values for all features and their summary, in relation to a baseline of predicted transformed data, are uniformly expressed in the same unit as the transformed data. This uniformity is particularly evident in local explanations where the baseline, alongside the feature responses in the force plot of an example player, aligns with the unit of the transformed data. This consistency in unit representation across different model outputs ensures a coherent understanding and interpretation of the prediction results, albeit with some complexity in the transformation process. 

Our study, while providing valuable insights, is subject to several limitations. Firstly, our current model is unable to accurately estimate the market value of superstar players, as this requires consideration of a broader range of societal and social factors that extend beyond the scope of our study and the data utilized. Consequently, future research should aim to include data pertaining to these additional factors and employ cross-disciplinary analytical methods to enhance the accuracy of superstar valuation.

Secondly, our data, sourced from public websites, encompasses a variety of features across different dimensions. However, this data set pales in comparison to the more comprehensive data captured by commercial providers, which can include over two hundred metrics per player per game. This disparity undoubtedly places limitations on the depth and breadth of insights our study can provide. In light of these limitations, future research endeavors should aim to apply our analytical pipeline to more robust datasets or to different problems at varying levels of analysis. These levels may include individual player assessments, team and league evaluations, and explorations in other fields where the prediction of value and the influence of various features are pertinent. Such expansions would not only enhance the applicability of our current model but also provide a more comprehensive understanding of value prediction in diverse contexts.

\section{Conclusion}
In conclusion, this study adeptly integrates advanced machine learning techniques and feature importance analysis, providing a nuanced understanding of player market value prediction within the realm of sports analytics. Our findings indicate that the Gradient Boosting Decision Tree (GBDT) model demonstrates a distinct advantage over other artificial intelligence algorithms in this context. The study identified key features that significantly influence player valuation, categorized into three dimensions: skill (encompassing Ball Control, Short Passing, Finishing, Interceptions, Dribbling, and Standing Tackle), fitness (including Sprint Speed and Acceleration), and cognitive (specifically, Reactions). The quantification of the impact of these features offers valuable insights for player evaluation, enhancing the precision and efficacy of market value assessments.

Looking ahead, there is considerable scope for future research to broaden this framework. By encompassing a more diverse range of features and applying the model in varied contexts, subsequent studies could achieve greater applicability and yield deeper insights. This expansion would not only reinforce the findings of the current study but also contribute to the ongoing advancement of machine learning applications in sports analytics and player valuation.

\bibliographystyle{unsrtnat}
\bibliography{references, cs, rommers}  





\clearpage
\begin{samepage}

\section*{Appendices}
\appendix
\section{Feature description}
\begin{table}[!h]
    \centering
    \caption{Feature description}  
    \begin{tabular}{lp{10cm}l}
    \hline
        feature name  & Description & data type \\ \hline
        name & ~ & string \\ 
        overall rating & ~ & integer \\ 
        potential & ~ & integer \\ 
        value & The estimated market value of a player & integer \\ 
        wage & ~ & integer \\ 
        Crossing & the ability to deliver accurate crosses & integer \\ 
        Finishing & the ability to score goals from various positions and situations. & integer \\ 
        Heading\_Accuracy & the ability to accurately direct headers towards the goal or teammates & integer \\ 
        Short\_Passing & the ability to accurately perform short passes to teammates & integer \\ 
        Volleys & the skill in striking the ball out of the air, often when attempting to score & integer \\ 
        Dribbling & the ability to control the ball and move past opponents. & integer \\ 
        Curve & the ability to apply spin and curve to passes and shots. & integer \\ 
        FK\_Accuracy & the skill in taking accurate free kicks. & integer \\ 
        Long\_Passing & the ability to accurately perform long passes to teammates. & integer \\ 
        Ball\_Control &  the skill in controlling the ball when receiving it or when dribbling. & integer \\ 
        Acceleration &  the ability to quickly reach their top speed. & integer \\ 
        Sprint\_Speed & the top running speed. & integer \\ 
        Agility & the ability to change direction quickly while maintaining control of the ball. & integer \\ 
        Reactions & the ability to quickly respond to in-game situations & integer \\ 
        Balance & the ability to maintain their balance when under physical pressure or when making quick movements. & integer \\ 
        Shot\_Power &  the ability to generate powerful shots on goal. & integer \\ 
        Jumping & the ability to jump high, which can be crucial in aerial duels or headers. & integer \\ 
        Stamina & the ability to maintain their performance level throughout the match without becoming fatigued. & integer \\ 
        Strength & the physical strength, which can help in challenges or holding off opponents. & integer \\ 
        Long\_Shots & the ability to accurately shoot from long distances. & integer \\ 
        Aggression & the tendency to engage in physical challenges and apply pressure on opponents. & integer \\ 
        Interceptions & the ability to read the game and intercept passes from the opposing team. & integer \\ 
        Positioning & the ability to be in the right place at the right time, both offensively and defensively. & integer \\ 
        Vision & the ability to see and make creative passes or plays during a match. & integer \\ 
        Penalties & the skill in taking penalty kicks. & integer \\ 
        Composure & the ability to remain calm and focused under pressure, which can impact their decision-making and performance. & integer \\ 
        Defensive\_Awareness & the ability to read the game defensively and position themselves effectively to make interceptions or tackles & integer \\ 
        Standing\_Tackle & the ability to effectively perform standing tackles to dispossess opponents without committing fouls. & integer \\ 
        Sliding\_Tackle & the skill in executing sliding tackles to win the ball from opponents while minimizing the risk of fouls or injuries. & integer \\ \hline
    \end{tabular}
    \label{table: feature description}
\end{table}
\end{samepage}

\clearpage
\section{Optimal hyperparameters for compared algorithms}
\begin{table}[!ht]
    \centering
    \caption{Optimal hyperparameters for compared algorithms}
    \begin{tabular}{ll}
    \hline
    \textbf{Model Name} & \textbf{Hyperparameters} \\
    \hline
    AdaBoost & learning\_rate=0.1, loss='linear', n\_estimators=150 \\
    GBDT & learning\_rate=0.1, max\_depth=3, n\_estimators=900 \\
    LightGBM & max\_depth=3, n\_estimators=900 \\
    Random Forest & max\_depth=15, min\_samples\_leaf=3, min\_samples\_split=3, n\_estimators=700 \\
    XGBoost & learning\_rate=0.3, max\_depth=3 \\
    CatBoost & depth=16, subsample=0.9 \\
    \hline
    \end{tabular}
    \label{tab:hyperparameters}
\end{table}

\end{document}